# Automatic Detection of Non-deverbal Event Nouns for Quick Lexicon Production

**Núria Bel**
IULA
Universitat Pompeu Fabra
nuria.bel@upf.edu

**Maria Coll**
IULA
Universitat Pompeu Fabra
maria.coll02@
campus.upf.edu

**Gabriela Resnik**
Universidad Nacional de
General Sarmiento and Universitat Pompeu Fabra
gresnik@ungs.edu.ar

## Abstract

In this work we present the results of experimental work on the development of lexical class-based lexica by automatic means. Our purpose is to assess the use of linguistic lexical-class based information as a feature selection methodology for the use of classifiers in quick lexical development. The results show that the approach can help reduce the human effort required in the development of language resources significantly.

## 1 Introduction

Although language independent, many linguistic technologies are inherently tied to the availability of particular language data (i.e. Language Resources, LR). The nature of these data is very much dependent on particular technologies and the applications where are used. Currently, most systems are using LR collected by hand that still do not cover all languages, or all possible application domains, or all possible information required by the many applications that are being proposed. Methods for the automatic and quick development of new LR have to be developed in order to guarantee a supply of the required data. Esuli and Sebastiani (2006) did a classification experiment for creating lexica for opinion mining, for instance, and the importance of lexical information for event extraction in Biomedical texts has been addressed in Fillmore et al. (2006). One way of producing such resources is to classify words into lexical classes via methods based on their morphosyntactic contexts of occurrence.

In the next three sections we report on an experiment on cue-based lexical classification for non-deverbal event nouns, that is, nouns such as 'party' or 'conflict', which refer to an event but cannot be identified by their morphology, as is the case with deverbal nouns such as 'construction'. The purpose of this experiment was, as already stated, to investigate methods for the rapid generation of an event nouns lexicon for two different languages, using a reduced quantity of available texts. Assuming that linguistic information can be provided by occurrence distribution, as is usually done in linguistic theory to motivate lexical classes (e.g. Grimshaw, 1990), cue information has been gathered from texts and used to train and test a Decision Tree-based classifier. We experimented with two different languages to test the potential coverage of the proposed technique in terms of its adaptation to different languages, and also used different types of corpora to test its adaptability to different domains and sizes.

## 2 Some properties of Non-Deverbal Event Nouns in Spanish and English.

We based our experiment on the work by Resnik (2004) who proposes a specific lexical class for Spanish event nouns like *accidente* ('accident') or *guerra* ('war') which cannot be identified by suffixes such as '-ción' ('-tion') or 'miento' ('-ment'), i.e. the morphological marks of deverbal derivation. Her proposal of creating a new class is motivated by the syntactic behaviour of these non-deverbal event nouns that differ significantly both from deverbal nominalizations and from non event nouns. This proposal differs significantly from work such as Grimshaw (1990).

In Grimshaw (1990) a significant difference is shown to exist between process and result nominals, evident in certain ambiguous nouns such as *building*, which can have a process reading – in a sentence like *The building of the access road took three* weeks (= 'process of building')– and a non-eventive or result reading –in a sentence like *The building collapsed* (= 'edifice'). These two types of nominals differ in many lexico-syntactic properties, such as the obligatory/optional internal argument realization, the manner of external argument realization, the determiner selection and their ability to control infinitival clauses. Simple event nouns such as *trip* share several syntactic properties with result nominals, although their lexical meaning is indeed similar to that of the process or complex event nouns. The main difference is the fact that result nominals and simple event nouns, contrary to complex event nominals, are not verb-like in the way they combine with their satellites (Grimshaw 1990). The similarity between result nominals and simple event nouns is accepted in Picallo's (1991, 1999) analysis of Catalan and Spanish nominalizations and in Alexiadou's (2001) work on nominalizations in Greek, English, Hebrew and other languages.

Although the similarities between non-deverbal event nouns like *accidente* and result nominals are undeniable, some evidence (Resnik, 2004 and 2009) has been found that non-deverbal event nouns cannot be assimilated to either result nominals or simple non event nouns like *tren* ('train'), in spite of their shared properties. In the next sections, we briefly present evidence that non-deverbal event nouns are a separate lexical class and that this evidence can be used for identifying the members of this class automatically, both in Spanish and in English. Our hypothesis is that whenever there is a lexical class motivated by a particular distributional behaviour, a learner can be trained to identify the members of this class. However, there are two main problems to lexical classification: noise and silence, as we will see in section 4.

Resnik (2004) shows that non-deverbal event nouns occur in a unique combination of syntactic patterns: they are basically similar to result nouns (and simple non event nouns) regarding the realization of argument structure, yet they pattern along process nominals regarding event structure, given that they accept the same range of aspectual adjuncts and quantifiers as these nouns and are selected as subjects by the same 'aspectual' verbs (*empezar*, 'to start'; *durar*, 'to last', etc.) (cf. section 3.2). As to other nominal properties, such as the mass/count distinction, the contexts show that non-deverbal event nouns are not quite like either of the two kinds of nominalizations, and they behave like simple non event nouns. The table below summarizes the lexico-syntactic properties of the different nouns described by Grimshaw (1990) with the addition of Resnik's proposed new one.

|  | NDV E N (*war*) | PR-N (*construction = event*) | RES-N (*construction = result. obj.*) | NEN (*map*) |
| --- | --- | --- | --- | --- |
| Obligatory internal argument | no | yes | no | No |
| External argument realization | genitive DP | PP_by | genitive DP | genitive DP |
| Subject of aspectual verbs (*begin, last..*) | yes | yes | no | no |
| Aspectual quantifier (*a period of*) | yes | yes | no | no |
| Complement of *during*, … | yes | yes | no | no |
| Count/mass (determiners, plural forms) | mass/count | mass | count | mass/count |

Table 1. Lexico-syntactic properties of English Non-Deverbal Event Nouns (NDV E N), Process Nouns (PR-N) and Result Nouns (RES-N) and Non Event Nouns (NEN).

## 3 Automatic Detection of Non-deverbal Event Nouns

We have referred to the singularities of non-deverbal event nouns as a lexical class in contrast with other event and non-event nouns. In our experiment, we have extracted the characteristics of the contexts where we hypothesize that members of this class occur and we have used them as variables to train an automatic learner that can rely on these features to automatically classify words into those which are indeed non-deverbal event nouns and those which are not. Because deverbal result nouns are easily identifiable by the nominal suffix they bear (for instance, '-tion' for English and '-ción' for Spanish), our experiment has been centered in separating non-deverbal event nouns like *guerra/war* from non event nouns like *tren/train*.

Some work related to our experiments can be found in the literature dealing with the identification of new events for broadcast news and semantic annotation of texts, which are two possible applications of automatic event detection (Allan et al. 1998 and Saurí et al. 2005, respectively, for example). For these systems, however, it would be difficult to find non-deverbal event nouns because of the absence of morphological suffixes, and therefore they could benefit from our learner.

### 3.1 Cue-based Lexical Information Acquisition

According to the linguistic tradition, words that can be inserted in the same contexts can be said to belong to the same class. Thus, lexical classes are linguistic generalizations drawn from the characteristics of the contexts where a number of words tend to appear. Consequently, one of the approaches to lexical acquisition proposes to classify words taking as input characteristics of the contexts where words of the same class occur. The idea behind this is that differences in the distribution of the contexts will separate words in different classes, e.g. the class of transitive verbs will show up in passive constructions, while the intransitive verbs will not. Thus, the whole set of occurrences (tokens) of a word are taken as cues for defining its class (the class of the type), either because the word is observed in a number of particular contexts or because it is not. Selected references for this approach are: Brent, 1993; Merlo and Stevenson, 2001; Baldwin and Bond, 2003; Baldwin, 2005; Joanis and Stevenson, 2003; Joanis et al. 2007.

Different supervised Machine Learning (ML) techniques have been applied to cue-based lexical acquisition. A learner is supplied with classified examples of words represented by numerical information about matched and not matched cues. The final exercise is to confirm that the data characterized by the linguistically motivated cues support indeed the division into the proposed classes. This was the approach taken by Merlo and Stevenson (2001), who worked with a Decision Tree and selected linguistic cues to classify English verbs into three classes: *unaccusative*, *unergative* and *object-drop*. Animacy of the subject*,* for instance, is a significant cue for the class of object dropping verbs, in contrast with verbs in *unergative* and *unaccusative* classes. Baldwin and Bond (2003) used a number of linguistic cues (i.e. co-occurence with particular determiners, number, etc.) to learn the countability of English nouns. Bel et al. (2007) proposed a number of cues for classifying nouns into different types according to a lexical typology. The need for using more general cues has also been pointed out, such as the part of speech tags of neighboring words (Baldwin, 2005), or general linguistic information as in Joanis et al. (2007), who used the frequency of filled syntactic positions or slots, tense and voice features, etc., to describe the whole system of English verbal classes.

### 3.2 Cues for the Detection of Non-deverbal Event Nouns in Spanish

As we have seen in section 2, non-deverbal event nouns can be identified by their occurrence in particular syntactic and lexical contexts of co-occurrence. We have used 11 cues for separating non-deverbal event nouns from non event nouns in Spanish. These cues are the following:

Cues 1-3. Nouns occurring in PPs headed by prepositions such as *durante* ('during'), *hasta el final de* ('until the end of'), *desde el principio de* ('from the beginning of'), and similar expressions are considered to be eventive. Thus, occurrence after one of such expressions will be indicative of an event noun.

Cues 4-8. Nouns occurring as external or internal arguments of verbs such as *ocurrir* ('occur'), *producir* ('produce' or 'occur', in the case of ergative variant *producirse*), *celebrar* ('celebrate'), and others with similar meanings, are also events. Note that we identify as 'external arguments' the nouns occurring immediately after the verb in particular constructions, as our *pos*- tagged text does not contain information about subjects (see below). In many cases it is the internal argument occurring in these contexts. These verbs tend to appear in 'presentative' constructions such as *Se produjo un accidente* ('An accident occurred'), with the pronoun *se* signalling the lack of external argument. Verbs like *ocurrir* appear in participial absolute constructions or with participial adjectives, which means they are unaccusatives.

Cue 9. The presence of temporal quantifying expressions such as *dos semanas de* ('two weeks of') or similar would indicate the eventive character of a noun occurring with it, as mentioned in section 2.

Cue 10. Non-deverbal event nouns will not be in Prepositional Phrases headed by locative prepositions such as *encima de* ('on top of') or *debajo de* ('under'). These cues are used as negative evidence for non-event deverbal nouns.

Cue 11. Non-deverbal event nouns do have an external argument that can also be realized as an adjective. The alternation of DP arguments with adjectives was then a good cue for detecting non-deverbal events, even when some other nouns may appear in this context as well. For instance: *fiesta nacional* ('national party') vs. *mapa nacional* ('national map').

### 3.3 Cues for the Detection of Non-Deverbal Event Nouns in English

As for Spanish, cues for English were meant to separate the newly proposed class of non-deverbal event nouns from non-event nouns if such a class exists as well.

Cues 1-3. Process nominals and non-deverbal event nouns can be identified by appearing as complements of aspectual PPs headed by prepositions like *during*, *after* and *before*, and complex prepositions such as *at the end of* and *at the beginning of*.

Cues 4 and 5. Non-deverbal nouns may occur as external or internal arguments of aspectual as well as occurrence verbs such as *initiate*, *take place*, *happen*, *begin*, and *occur*. Those arguments are identified either as subjects of active or passive sentences, depending on the verb, i.e. *the therapy was initiated* and *the conflict took place*.

Cue 6. Likewise, nouns occurring in expressions such as *frequency of*, *occurrence of* and *period of* would probably be event nouns, i.e. *the frequency of droughts*.

Cue 7 and 8. Event nouns may as well appear as objects of aspectual and time-related verbs, such as in *have begun a campaign* or *have carried out a campaign*.

Cues 10 and 11. They are intended to register event nouns whose external argument, although optional, is realized as a genitive complement, e.g. *enzyme's loss*, even though this cue is shared with other types of nouns. Following the characterization suggested for Spanish, we also tried external arguments realized as adjectives in cue 11, as in *Napoleonic war*, but we found empirical evidence that it is not useful.

Cues 12-16. Finally, as in the experiment for Spanish, we have also included evidence that is more common for non-event nouns, that is, we have used negative evidence to tackle the problem of sparse data or silence discussed in the next section. It is considered a negative cue for a noun to be preceded by an indefinite determiner, to be in a PP headed by a locative preposition, and to be followed by the prepositions *by* or *of*, as a PP headed by one these prepositions could be an external argument and, as it has been noted above, the external argument of event nouns tends to be realized as a genitive complement (as in *John's trip/party*).

In the selection of these cues, we have concentrated on those that separate the class of non-deverbal event nouns from the class formed by simple non event nouns like *train*, where no particular deverbal suffix can assist their detection. If it is the case that these are really cues for detecting non-deverbal event nouns, the learner should confirm it by classifying non-deverbal event nouns correctly, separating them from other types of nouns.

## 4 Experiment and results

For our experiments we have used Regular Expressions to implement the patterns just mentioned, which look for the intended cues in a part-of-speech tagged corpus. We have used a corpus of 21M tokens from two Spanish newspapers (*El País* and *La Vanguardia*), and an English technical corpus made of texts dealing with varying subject matter (Economy, Medicine, Computer science and Environmental issues), of about 3.2M tokens. Both Spanish and English corpora are part of the Technical Corpus of IULA at the UPF (CT-IULA, Cabré et al. 2006). The positive or negative results of the n-pattern checking in all the occurrences of a word are stored in an n-dimension vector. Thus, a single vector summarizes all the occurrences of a word (the type) by encoding how many times each cue has been observed. Zero values, i.e. no matching, are also registered.

We used a Decision Tree (DT) classifier in the Weka (Witten and Frank, 2005) implementation of pruned C4.5 decision tree (Quinlan,

1993). The DT performs a general to specific search in a feature space, selecting the most informative attributes for a tree structure as the search proceeds. The goal is to select the minimal set of attributes that efficiently partitions the feature space into classes of observations and assemble them into a tree. During the experiment, we tuned the list of cues actually used in the classification task, because some of them turned out to be useless, as they did not show up even once in the corpus. This was especially true for the English corpus with cues 5, 11 and 12. Note that the English corpus is only 3.2 million words.

In the experiment we used a 10-fold cross-validation testing using manually annotated gold-standard files made of 99 non-event and 100 non-deverbal event nouns for Spanish and 93 non event and 74 non-deverbal event nouns for English[1]. In this first experiment, we decided to use mostly non-deverbal non event nouns such as *map*, because detecting result nouns like *construction* is easy enough, due to the deverbal suffix. However, for the English experiment, and because of the scarcity of non-deverbal nouns occurrences, we had to randomly select some deverbals that were not recognized by the suffix.

The results of our experiment gave a total accuracy of 80% for Spanish and 79.6% for English, which leads to think that corpus size is not a determinant factor and that this method can be used for addressing different languages, provided a good characterization of the lexical class in terms of particular occurrence distributions is achieved. Yet, although the accuracy of both English and Spanish test sets is similar, we will see later on that the size of the corpus does indeed affect the results.

An analysis of the errors shows that they can be classified in two groups: errors due to noise, and errors due to silence.

(i) Noise. In his seminal work, Brent (1993) already pointed out that "the cues occur in contexts that were not aimed at". Noise can be due to errors in processing the text, because we had only used low-level analysis tools. For instance, in "during the first world war" our RE cannot detect that "world" is not the head of the Noun Phrase. Brent's hypothesis, followed by most authors afterwards, is that noise can be eliminated by statistical methods because of its low frequency. However, the fact is that in our test set significant information is as sparse as noise, and the DT cannot correctly handle this. In our data sets, most of the false positives are due to noise.

(ii) Silence. Some nouns appear only once or twice in the corpus and do not show up in any of the sought contexts (for instance, *terremoto*, 'earthquake', in Spanish press). Moreover, this is independent of the size of the corpus, because the Zipfian distribution of tokens allows us to predict that there will always be low-frequency nouns. Low frequency words produce non informative vectors, with only zero-valued cues, and our classifier tends to classify non-informative vectors as non-event nouns, because most of the cues have been issued to identify event nouns. This was the main reason to introduce negative contexts as well as positive ones, as we mentioned in section 3.

However, these systematic sources of error can be taken as an advantage when assessing the usability of the resulting resources. Having about 80% of accuracy would not be enough to ensure the proper functioning of the application in which the resource is going to be used. So, in order to gain precision, we decided to separate the set of words that could be safely taken as correctly classified. Thus, we had used the confidence, i.e. probability of the classification de-cisions to assess which are below a reasonable level of confidence.

---

[1] **Positive:** accident, assembly, audience, battle, boycott, campaign, catastrophe, ceremony, cold, collapse, conference, conflict, course, crime, crisis, cycle, cyclone, change, choice, decline, disease, disaster, drought, earthquake, epidemic, event, excursion, fair, famine, feast, festival, fever, fight, fire, flight, flood, growth, holiday, hurricane, impact, incident, increase, injury, interview, journey, lecture, loss, meal, measurement, meiosis, marriage, mitosis, monsoon, period, process, program, quake, response, seminar, snowstorm, speech, storm, strike, struggle, summit, symposium, therapy, tour, treaty, trial, trip, vacation, war. **Negative:** agency, airport, animal, architecture, bag, battery, bird, bridge, bus, canal, circle, city, climate, community, company, computer, constitution, country, creature, customer, chain, chair, channel, characteristic, child, defence, director, drug, economy, ecosystem, energy, face, family, firm, folder, food, grade, grant, group, health, hope, hospital, house, illusion, information, intelligence, internet, island, malaria, mammal, map, market, mountain, nation, nature, ocean, office, organism, pencil, people, perspective, phone, pipe, plan, plant, profile, profit, reserve, river, role, satellite, school, sea, shape, source, space, star, statistics, store, technology, television, temperature, theme, theory, tree, medicine, tube, university, visa, visitor, water, weather, window, world.

In the Spanish test set, for instance, precision of the positive classification, i.e. the percentage of words correctly classified as event nouns, raises from 0.82 to 0.95 when only instances of classification with a confidence of more than 0.8 are selected. In the figure below, we can see the precision curve for the Spanish test set.

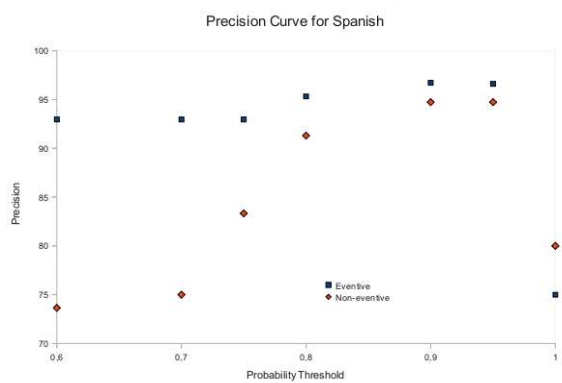

Figure 1: Precision curve for the Spanish test set.

In general, precision is higher when confidence is higher, except for complete confidence, 1, as we will explain later with the English case. This general behavior could be interpreted as a guarantee that there is a significant number of classified nouns (87 out of 199 for the Spanish test set with a threshold of 0.8 confidence) that need not to be manually reviewed, i.e. a 43% of the automatically acquired lexica can safely be considered correct. From figure 1, we can also see that the classifier is consistently identifying the class of non-deverbal event nouns even with a lower threshold. However, the resulting non-event noun set contains a significant number of errors. From the point of view of the usability, we could also say that only those words that are classified as non-event nouns must be revised.

Figure 2 for English test set shows a different behavior, which can only be justified because of the difference in corpus size. A small corpus increases the significance of silence errors. Fewer examples give less information to the classifier, which still makes the right decisions but with less confidence in general. However, for the extreme cases, for instance the case of 7 word vectors with only zero-values, the confidence is very high, that is 1, but the decisions are wrong. These cases of mostly zero values are wrongly considered to be non-events. This is the reason for the low precision of very confident decisions in English, i.e. sparse data and its consequence, silence.

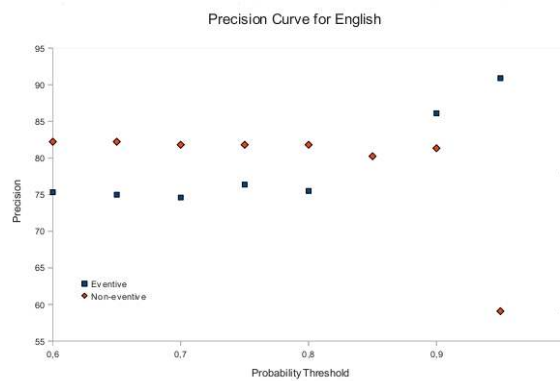

Figure 2: Precision curve for the English test set.

## 5 Conclusions

In this paper we have proposed the use of lexical classification methods based on differences in the distributional behavior of word classes for the quick production of lexica containing the information required by particular applications. We have dealt with non-deverbal event nouns, which cannot be easily recognized by any suffixes, and we have carried out a classification experiment, which consisted in training a DT with the information used in the linguistic literature to justify the existence of this class. The results of the classifier, close to 80% accuracy in two different languages and with different size and types of source corpora, show the validity of this very simple approach, which can be decisive in the production of lexica with the knowledge required by different technologies and applications in a time-efficient way. From the point of view of usability, this approach can be said to reduce the amount of work in more than a 40%.

## Acknowledgements

We want to thank Muntsa Padró for her valuable contribution. This work was partially supported by the PANACEA project (EU-7FP-ITC-248064).